\documentclass[letterpaper, 10 pt, conference]{ieeeconf}  

\IEEEoverridecommandlockouts                              

\usepackage{graphics} 
\usepackage{epsfig} 
\usepackage{amsmath} 
\usepackage{amssymb}  
\usepackage{siunitx}
\usepackage{booktabs}
\usepackage{multicol}
\usepackage{multirow}
\usepackage{xcolor}
\usepackage[noadjust]{cite}
\title{\LARGE \bf
Characterization of Real-time Haptic Feedback from Multimodal\\ Neural Network-based Force Estimates during Teleoperation
}

\author{Zonghe Chua$^{1}$ and Allison M. Okamura$^{1}$
\thanks{*This work was supported by a Stanford Bio-X Fellowship.}
\thanks{$^{1}$Z. Chua and A. M. Okamura are with the Mechanical Engineering Department, Stanford University, CA 94305, USA.
{\tt\small chuazh@stanford.edu, aokamura@stanford.edu}}%
}

\begin{document}

\bstctlcite{IEEEexample:BSTcontrol} 
\newcommand{\hlight}[1]{\textcolor{red}{#1}}

\maketitle
\thispagestyle{empty}
\pagestyle{empty}

\begin{abstract}

Force estimation using neural networks is a promising approach to enable haptic feedback in minimally invasive surgical robots without end-effector force sensors. Various network architectures have been proposed, but none have been tested in real time with surgical-like manipulations. Thus, questions remain about the real-time transparency and stability of force feedback from neural network-based force estimates. We characterize the real-time impedance transparency and stability of force feedback rendered on a da Vinci Research Kit teleoperated surgical robot using neural networks with vision-only, state-only, and state and vision inputs. Networks were trained on an existing dataset of teleoperated manipulations without force feedback. To measure real-time stability and transparency during teleoperation with force feedback to the operator, we modeled a one-degree-of-freedom human and surgeon-side manipulandum that moved the patient-side robot to perform manipulations on silicone artificial tissue over various robot and camera configurations, and tools. We found that the networks using state inputs displayed more transparent impedance than a vision-only network. However, state-based networks displayed large instability when used to provide force feedback during lateral manipulation of the silicone. In contrast, the vision-only network showed consistent stability in all the evaluated directions. We confirmed the performance of the vision-only network for real-time force feedback in a demonstration with a human teleoperator.
\end{abstract}


\section{INTRODUCTION}

Safe tissue handling is an important and often evaluated component of human skill for robot-assisted minimally invasive surgery (RMIS) \cite{goh2012global}\cite{martin1997objective}. However, it is thought to be a difficult skill to develop due to the lack of haptic feedback in teleoperated RMIS platforms. Displaying kinesthetic force feedback to the surgeon is one way to overcome this limitation. This typically requires three-degree-of-freedom force sensing at the tip of the robot end-effector with forces displayed to the teleoperator at high update rates. It is difficult to achieve this cost-effectively in RMIS due to the small size of the surgical manipulators and requirements for biocompatibility and sterilizability \cite{Enayati2016}.

To circumvent direct force sensing, force estimation through physics-based dynamic models that use robot kinematic and dynamic state inputs have been explored \cite{fontanelli2017modelling,Pique2019dynamic,Wang2019dynamic}, though they do not accurately estimate external joint torques in the most distal degrees of freedom, especially outside the range of joint angles and velocities used to fit their parameters. Vision-based methods that try to estimate interaction force from environmental deformation have also been explored through the use of stereo image reconstruction and finite-element mesh fitting, but these models are computationally expensive and highly sensitive to the definition of the mesh anchors \cite{Haouchine2018FEA}.

Neural networks are a promising approach to estimate force, replacing the need for explicit model specification with the need for more data. This approach has been used to model the dynamics of the robot in free space to predict external joint torques during environment interaction using only robot state inputs through supervised \cite{Tran2020} or self-supervised learning \cite{Yilmaz2020dynamic}. Other methods attempt to use both vision- and state-based inputs to estimate forces, with many architectures also incorporating sequential temporal inputs either through recurrent neural networks \cite{aviles2016towards,Marban2018,Marban2019rnn,LeeDaVinciNeuralNet} or a transformer network \cite{Shin2019transformer}. Our prior work has shown that static image and robot state-based models that do not encode temporal history can also estimate forces with similar accuracy, with the added benefit of faster computation time \cite{ChuaNeuralForce}. Deep learning approaches provided good offline accuracy relative to physics-based dynamic models, making them promising for applications such as offline surgical skill evaluation. With the exception of Tran et al. \cite{Tran2020} who conducted a pilot stiffness discrimination study with force feedback from a neural network , these methods have not been evaluated in real-time teleoperation. Thus, questions remain about their real-time performance in terms of impedance transparency and human-in-the-loop stability.


In this work, we characterized the impedance transparency and stability of force estimation neural networks that use either robot state-only, vision-only, or multimodal (vision and robot state) inputs from a single time point to output an end-effector force estimate in the three Cartesian directions. We developed a one-degree-of-freedom model of the human teleoperator and surgeon-side manipulandum (SSM) whose movements command the patient-side manipulator (PSM) to perform manipulations on physical silicone mock tissue in each of the Cartesian directions, thus mimicking bilateral teleoperation. These manipulations were performed under different robot and camera configurations as well as with different surgical tools. We recorded the estimated forces, ground truth forces, and end-effector displacements to derive measures of impedance transparency. To quantify stability, we introduce a passivity-based stabilizing effort metric.  For comparison, we used force estimates from a physics-based dynamic model and the ground truth interaction force from a force sensor placed underneath the artificial tissue.

\section{METHODS}

\begin{figure}[!t]
\centering
\vspace{1.7mm}
\includegraphics[width=\linewidth]{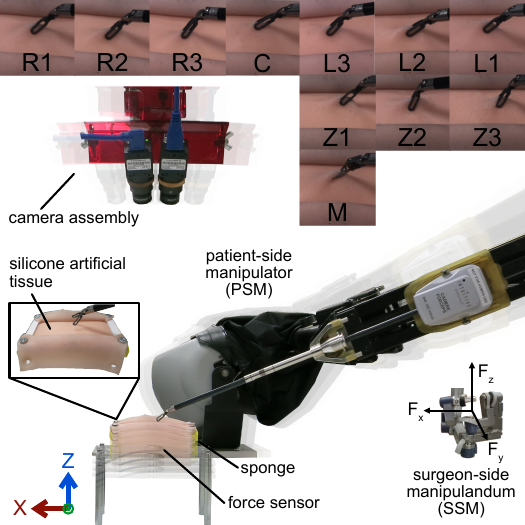}
\caption{Experimental set up comprising a da Vinci Research Kit (patient-side manipulator and surgeon-side manipulandum) and a silicone artificial tissue mounted on a stage with a force sensor placed underneath. The center (C) configuration is shown with the other configurations (R1-3, L1-3 and Z1-3) translucently overlaid. The image tiles displayed at the top are the cropped camera images from the left camera for each configuration. In the M condition, the patient-side manipulator is equipped with Maryland Bipolar forceps. In the other conditions it is equipped with Cadiere forceps.}
\label{teleop}
\end{figure}

\subsection{Hardware Setup}

A da Vinci Research Kit (dVRK) right PSM \cite{kazanzides2014dvrk}, a stereo camera assembly, and a custom artificial tissue mounting platform (Fig.\,\ref{teleop}) were used for all experiments. In addition, the dVRK right SSM (Fig.\,\ref{teleop} inset) was also used when actual teleoperation was required. The stereo camera assembly consisted of two cameras (Flea3, FLIR Systems, Wilsonville, OR, USA) that captured 960$\times$540 images at 30\,Hz. On this platform, a piece of artificial tissue (Professional Skin Pad Mk1, Limbs \& Things, Savannah, GA, USA) measuring 125$\times$72$\times$3\,mm, with the silicone layer separated from the sponge layer, was clamped to a rigid base with a 6-axis force-torque sensor (Nano17, ATI Industrial Automation, Apex, NC, USA) mounted beneath. To ensure a tissue-like border around the manipulated artificial tissue, the same silicone material was clamped to the top and bottom border of the tissue. A Cadiere forceps tool was the default tool attached to the PSM. The camera was oriented such that the left camera field of view was always centered on the middle of the tissue and its field of view cropped to 300$\times$300 before being resized to 224$\times$224 for input to the vision-based networks. 

To collect neural network training data and perform our force feedback experiments over different robot and camera configurations, the robot base and camera could be translated in the positive and negative x direction. The center (C) position was where the robot base was positioned +14.5\,cm in the x direction relative to the center line of the manipulation stage with the camera assembly centered on the stage. Relative to the C configuration, we defined three levels of translations of the robot base and camera assembly of 1.5cm, 3.0cm, and 4.0cm in the positive (R1-3) and negative (L1-3) x direction. The R and L definitions thus describe the effective translation of the manipulation stage relative to the camera assembly and robot base. To capture vertical translations of the stage, we defined three levels of translations of the manipulation stage of 0.635\,cm, 1.270\,cm, and 1.905\,cm in the negative z direction (Z1-3).

Control of the robot, sensor output processing, and dynamics simulation were split between two computers networked through Robot Operating System (ROS). The first computer handled the dVRK low-level control, robot state sensing, and force sensing processes at 1000\,Hz. The second computer ran the image processing, dynamics simulation, force estimation, and high-level arm trajectory command processes.

\subsection{Force Estimation Model Architectures and Training}

\begin{table}[!t]
\vspace{1.5mm}
\centering
\caption{Number of Examples Used in Training and Validation}
\label{table:trainsplit}
\begin{tabular}{@{}ccc@{}} 
\toprule
Configuration & Training & Validation  \\ 
\midrule
Center             & 14020 & 7036        \\
Right 2            & 14105 & 7055        \\
Left 2            & 14072 & 7065        \\
Z2            & 6977  & 3503        \\
\bottomrule
\end{tabular}
\end{table}

The force estimation neural networks used either state-only (S), vision-only (V), or vision and state (VS) inputs as described in \cite{ChuaNeuralForce}. In all networks, the final output was a force estimate in the three Cartesian dimensions. Vision inputs were RGB monocular images of size 224$\times$224$\times$3, while state inputs were 54-dimensional vectors composed of robot joint positions, velocities, and torques, and Cartesian position, velocity, and force. All image inputs were normalized by the ImageNet \cite{deng2009imagenet} mean and standard deviation. 

The S network was a fully connected network with 6 layers of sizes 500, 1000$\times$3, 500, and 50, respectively. The V network was a ResNet50 \cite{he2016resnet} architecture initialized on pre-trained ImageNet weights. During training, only the residual and fully connected layers were fine-tuned on our training data. The VS network used the same ResNet50 front-end but with a 30-dimensional output that was concatenated with the robot state input before passing it through three fully connected layers of size 84, 180, and 50, respectively. 

Each type of neural network was fit on data of teleoperated manipulations on artificial tissue performed with no force feedback to the human operator. These manipulations were done in the R2, C, L2, and Z2 configurations with the image and robot state data recorded at 30\,Hz to create training and validation data sets as summarized in Table \ref{table:trainsplit}. The networks were optimized with mean squared error loss using Adam \cite{KingmaAdam} with learning rates of 0.001 for the S and VS models and 0.0001 for the V model, and L1 regularization of 0.001 over 100 epochs. After each epoch, performance on the validation set was evaluated. The parameters that resulted in the best performance over the 100 epochs were used in the final models. Image augmentation was not used as it had a negligible effect on the performance of the network. During inference, the state inputs were normalized by the mean and standard deviation of the training data set.

A physics-based dynamic model \cite{Wang2019dynamic}, henceforth denoted as D, was used as a baseline force estimation model. We performed parameter estimation using the methods from \cite{Wang2019dynamic} and subtracted the estimated joint torques from the measured joint torques to obtain the interaction joint torques. The interaction torques were multiplied with the pseudoinverse of the Jacobian transpose to arrive at the estimated interaction forces at the end-effector. In all experiments, the estimated forces from D were low-pass filtered at a cutoff of 1\,Hz. This low cutoff was required to reduce the force oscillations that induced large movement oscillations during the closed-loop force feedback manipulation experiment. Without filtering, the manipulated silicone would slip out of grasp resulting in failed experiments. For ground truth comparison, we used forces measured through a force sensor sampled at 1000\,Hz. This method will be denoted as FS.

\subsection{One-Degree-of-Freedom Closed-loop Force Feedback Manipulation with Modeled Teleoperation Dynamics}
\label{methodsClosedLoop}
\begin{figure}[!t]
\centering
\includegraphics[width=0.9\linewidth]{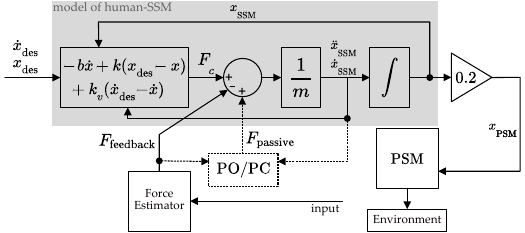}
\caption{Block diagrams for the one-degree-of-freedom closed-loop force feedback manipulation (Sec.\,\ref{methodsClosedLoop}). The force estimator block represents the different force estimation methods. The input to this block varies depending on the type of method. The PO/PC block is a passivity observer and controller module that is only included in experiments where passivity is being measured.}
\label{controlflow}
\end{figure}

Human teleoperators have natural variability in their movements that make it difficult for them to consistently replicate trajectories even under the same conditions. This makes quantifying and comparing transparency and stability behavior difficult across force estimation methods. In addition, the coupling of dynamics in three dimensions makes it hard to identify the contributions of each estimated force direction to transparency and stability. Thus, we defined a combined model of the human and the SSM in one degree of freedom. This was done by approximating a combined human-SSM model in each of the Cartesian directions as a one-dimensional mass-spring-damper system such that
\begin{equation}
    F_{c}-F_{\text{feedback}}= m\ddot{x} 
    \label{handmsd}
\end{equation}
where the combined mass of the human hand and SSM is $m = 0.750\, \text{kg} $, and $F_{\text{feedback}}$ is the force feedback from the force estimation method. $F_{\text{c}}$ is the human tracking force defined as
\begin{equation}
    F_{\text{c}} = \begin{cases} -b\dot{x} + k(x_{\text{des}}-x) + 
    k_v(\dot{x}_{\text{des}}-\dot{x}) & \text{if moving} \\
    -b\dot{x} + k(x_{\text{des}}-x) & \text{otherwise}
    \end{cases}
    \label{velocitytrack}
\end{equation}
where $x_{\text{des}}$ and $\dot{x}_{\text{des}}$ are desired displacement and velocity commands. The damping term $b = 6.45\text{\,Ns\,m}^{-1}$, and stiffness term $k=135\,\text{N\,m}^{-1}$ were set according to the values found from human arm dynamics characterization in \cite{vanbeeke2015stiffness}. The term $k_v=20\text{\,Ns\,m}^{-1}$ was tuned empirically to ensure good trajectory tracking. In Eqn.\,\ref{velocitytrack}, ``moving" describes the periods of time when the modeled human-SSM was displacing the material to a target position, as opposed to just trying to hold a static position. The parameters for Eqns.\,\ref{handmsd} and \ref{velocitytrack} were empirically confirmed to produce outputs that approximated actual human teleoperation. The arm dynamics were simulated at 500\,Hz with the resultant velocity integrated into a desired displacement, scaled by a factor of 0.2, and commanded to the PSM at a rate of 500\,Hz. Force feedback was provided based on estimates from each force estimation method with V and VS run at a rate of 60\,Hz, S at 500\,Hz, D at 400\,Hz, and FS at 1000\,Hz. Fig.\,\ref{controlflow} shows a block diagram of this system.

\begin{figure}[!t]
\centering
\vspace{1.6mm}
\includegraphics[width=0.9\linewidth]{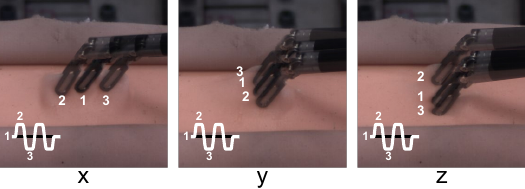}
\caption{Directions of movement of the patient-side manipulator for the one-degree-of-freedom closed-loop force feedback manipulation using a model of the human and surgeon-side manipulator in the x, y, and z Cartesian directions. Insets show relative movement timings and amplitude.}
\label{movements}
\end{figure}


Before each trajectory sequence was commanded, the PSM was initialized by first moving to grasp the tissue before translating to a position that zeroed out any forces induced from the grasp. For manipulations in the x and y directions, the material was further pre-tensioned in the z direction by 1\,N. Then a sequence of minimum jerk trajectories were commanded to the human plant, first by +75\,mm in 1\,s, then by -150\,mm in 2\,s, +150\,mm in 2\,s, \mbox{-150\,mm} in 2\,s before returning to the pre-tensioned position by moving +75\,mm in 1\,s. Between each movement, the position was held for 2\,s. These movements, when scaled by the 0.2 scaling factor, are equivalent to ideal displacements of the PSM by 15\,mm or 30\,mm (Fig.\,\ref{movements}). For the z direction manipulations, there was no pre-tensioning. This allowed the commanded manipulations to cover both compression and tension regimes. Instead, the same minimum jerk trajectories were commanded with the final movement returning the PSM to the initial force-neutral position (Fig.\,\ref{movements}).

To quantify impedance transparency during the manipulations, we fitted a Kelvin-Voigt viscoelastic model, defined as
\begin{equation}
    F= k_ix + b_i\dot{x} + c \qquad i = \{t,p\} ,\, k\geq 0, \, b \geq 0
    \label{stiffness_eq}
\end{equation}
where $k$ is a stiffness term, $b$ is a damping term, and $c$ is an additional offset term that accounts for any pre-tensioning error during initialization or systematic estimation error at zero displacement. The index $t$ denotes the parameters in tension (all directions) and $p$ the parameters in palpation (z direction only). Parameters were fit for each performed trajectory. While more complex linear and non-linear models can better represent the dynamics of the artificial tissue material, the Kelvin-Voigt model is interpretable and can well describe the dynamics of the rubber-like materials in the range of our applied strains.

We calculated an additional metric, the root mean squared error (RMSE) of the estimated force with respect to the ground truth force.
To quantify the stability of the system with closed-loop force feedback, we adopted a passivity-based approach. The complex dynamics of the robot, coupled with the non-dynamical multimodal input-output formulation of the neural networks, and their black-box nature, makes Lyapunov analysis intractable. 

The passivity-based analysis was performed by implementing a passivity observer and controller on the SSM-side and measuring the root mean square (RMS) control effort produced by the passivity controller in its attempt to passivate the system. A windowed  passivity observer was implemented such that the windowed energy observed at the SSM on the $\text{N}^{\text{th}}$ time step is
\begin{equation}
    E_{\text{win}}(N) = \sum_{n=N-k+1}^N F_{\text{feedback}}(n)\dot{x}(n)\Delta t
\end{equation}
where $k=10$ is the sampling window size and $\Delta t$ the period between the $n^{\text{th}}$ and the $(n-1)^{\text{th}}$ measurement \cite{Jorda2017window}. $k$ was tuned empirically so that the controller could react quickly to negative energy production as opposed to storing energy and only going negative after a period of negative energy was produced. A passivity controller \cite{Balachandran2017passivity} was used to generate the passivating control force using the control rule
\begin{equation}
    F_{\text{passive}}(n) = \begin{cases}
    -\dot{x}\frac{E_{\text{win}}(n)}{\dot{x}^2\Delta t} & \text{if } E_{\text{win}}(n)<0 \\
    0 & \text{otherwise}
    \end{cases}
\end{equation}
where the variable damping term $\frac{E_{\text{win}}(n)}{\dot{x}^2\Delta t}$ was limited to a maximum value of 250\,Ns\,m\textsuperscript{-1}. The combined passivity controller and observer module was run at 1000\,Hz.

For each type of force estimation method with and without the passivity module, the trajectory sequence for each Cartesian direction was performed three times on three different silicone samples with the order of the estimation method and passivity condition (with and without the passivity module) randomized. The displacement of the robot end-effector was measured via its joint encoders.

\subsection{Demonstration of Teleoperation with Force Feedback}

To demonstrate the capability of neural network-based force estimation for closed-loop force feedback in teleoperation, we chose to prioritize a network that had the best stability as opposed to transparency. A human teleoperated the SSM with force feedback and performed a series of slow and fast palpations and retractions on the artificial tissue in the three Cartesian degrees of freedom over 52\,s. 


\begin{table}[!t]
\vspace{1.5mm}
\centering
\caption{Mean and Standard Deviation of Fit Parameters over All Configurations}
\label{table:stiff}
\resizebox{\columnwidth}{!}{
\begin{tabular}{@{}cc r@{\,}c@{\,}l r@{\,}c@{\,}l r@{\,}c@{\,}l@{}} 
\toprule
Condition & Configuration & \multicolumn{3}{c}{k (Nm\textsuperscript{-1})}   & \multicolumn{3}{c}{b (Nsm\textsuperscript{-1})} & \multicolumn{3}{c}{c (N)}            \\ 
\midrule
\multirow{4}{*}{S~}                                 & x             & 229.14&$\pm$&22.39         & 39.76&$\pm$&11.23           & -0.28&$\pm$& 0.14  \\
                                                    & y             & 259.87&$\pm$&21.63         & 31.66&$\pm$&2.23            & 0.07&$\pm$& 0.12   \\
                                                    & z-tension     & 142.49&$\pm$&25.02         & 13.75&$\pm$& 4.66            & -0.27&$\pm$& 0.32  \\
                                                    & z-palpation   & 380.14&$\pm$&51.73         & 24.59&$\pm$& 5.65            & 0.01&$\pm$& 0.43   \\
\\
\multirow{4}{*}{V~}                                 & x             & 188.19&$\pm$&22.81         & 0.00&$\pm$& 0.00             & -0.03&$\pm$& 0.21  \\
                                                    & y             & 88.30&$\pm$&15.30          & 0.00&$\pm$& 0.00             & -0.02&$\pm$& 0.22  \\
                                                    & z-tension     & 79.07&$\pm$&20.66          & 0.00&$\pm$& 0.00             & -0.11&$\pm$& 0.21  \\
                                                    & z-palpation   & 200.63&$\pm$&48.89         & 0.00&$\pm$& 0.00             & 0.00&$\pm$& 0.24   \\
\\
\multirow{4}{*}{VS}                                 & x             & 241.65&$\pm$&29.80         & 25.73&$\pm$& 8.39            & 0.03&$\pm$& 0.26   \\
                                                    & y             & 242.44&$\pm$&39.32         & 23.88&$\pm$& 4.03            & 0.27&$\pm$& 0.17   \\
                                                    & z-tension     & 155.48&$\pm$&46.02         & 9.54&$\pm$& 3.32             & 0.25&$\pm$& 0.87   \\
                                                    & z-palpation   & 367.88&$\pm$&53.69         & 17.66&$\pm$& 5.57            & 0.61&$\pm$& 1.00   \\
\\
\multirow{4}{*}{D}                                  & x             & 376.79&$\pm$&32.10         & 0.66&$\pm$& 1.79             & -0.05&$\pm$& 0.32  \\
                                                    & y             & 513.58&$\pm$&51.87         & 0.00&$\pm$& 0.00             & -0.15&$\pm$& 0.26  \\
                                                    & z-tension     & 282.38&$\pm$&30.89         & 0.20&$\pm$& 1.13             & -0.24&$\pm$& 0.42  \\
                                                    & z-palpation   & 499.47&$\pm$&42.59         & 0.00&$\pm$& 0.00             & 0.03&$\pm$& 0.41   \\
\\
\multirow{4}{*}{FS}                                 & x             & 302.54&$\pm$&32.32         & 27.02&$\pm$& 3.75            & -0.04&$\pm$& 0.07  \\
                                                    & y             & 314.13&$\pm$&38.77         & 30.65&$\pm$& 4.97            & 0.14&$\pm$& 0.11   \\
                                                    & z-tension     & 192.99&$\pm$&26.65         & 10.60&$\pm$& 1.71            & -0.24&$\pm$& 0.04  \\
                                                    & z-palpation   & 334.82&$\pm$&26.94         & 23.17&$\pm$& 2.75            & -0.11&$\pm$& 0.03  \\
\bottomrule
\end{tabular}
}
\end{table}

\section{RESULTS AND DISCUSSION}

\begin{figure*}[!t]
\centering
\vspace{1.5mm}
\includegraphics[width=\linewidth]{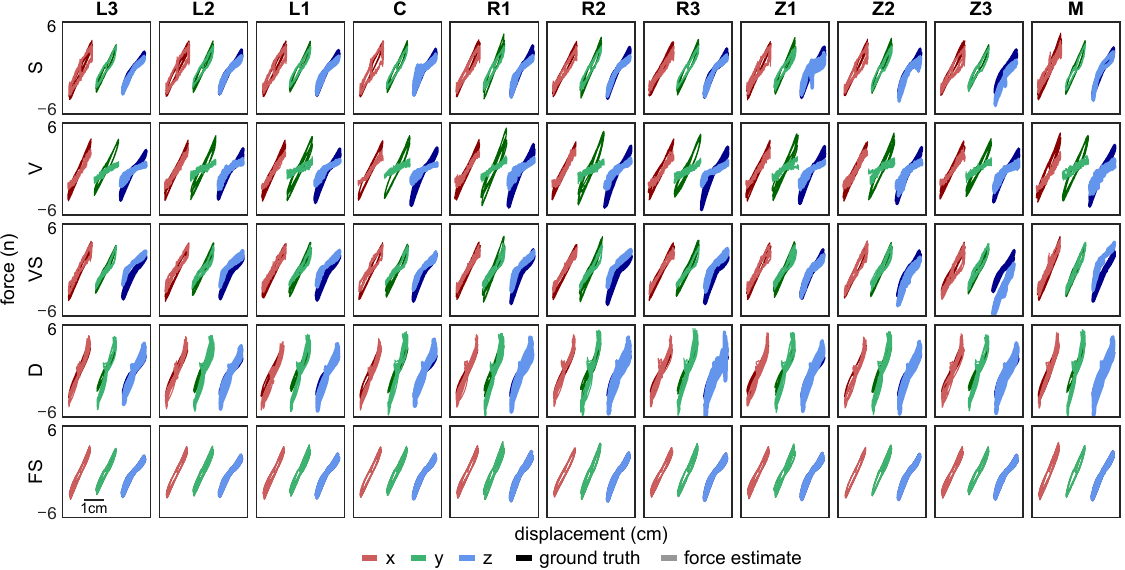}
\caption{Force-displacement plots for all robot and camera configuration in the three Cartesian directions from closed-loop force feedback from each force estimation method with all trials overlaid. Each Cartesian direction is offset in the horizontal displacement axis from the previous by 2.5cm.}
\label{stiffplots}
\end{figure*}

\begin{figure}[!t]
\centering
\includegraphics[width=\linewidth]{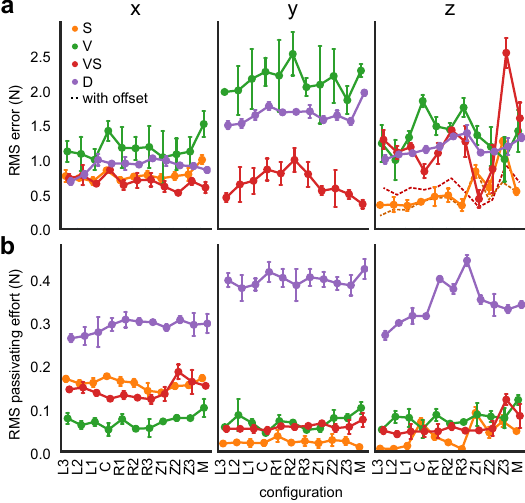}
\caption{(a) Root mean squared (RMS) error of the estimated force feedback compared to the ground truth force measurement. Dashed line represents the RMS error when the force bias of the estimation is offset to zero. (b) RMS passivating control effort during periods when the target position was held by the human-manipulandum model for each Cartesian direction, averaged over three different materials. Error bars denote $\pm$1 standard deviation.}
\label{summary_stats}
\end{figure}

\begin{figure*}[!t]
\centering
\vspace{1.5mm}
\includegraphics[width=\linewidth]{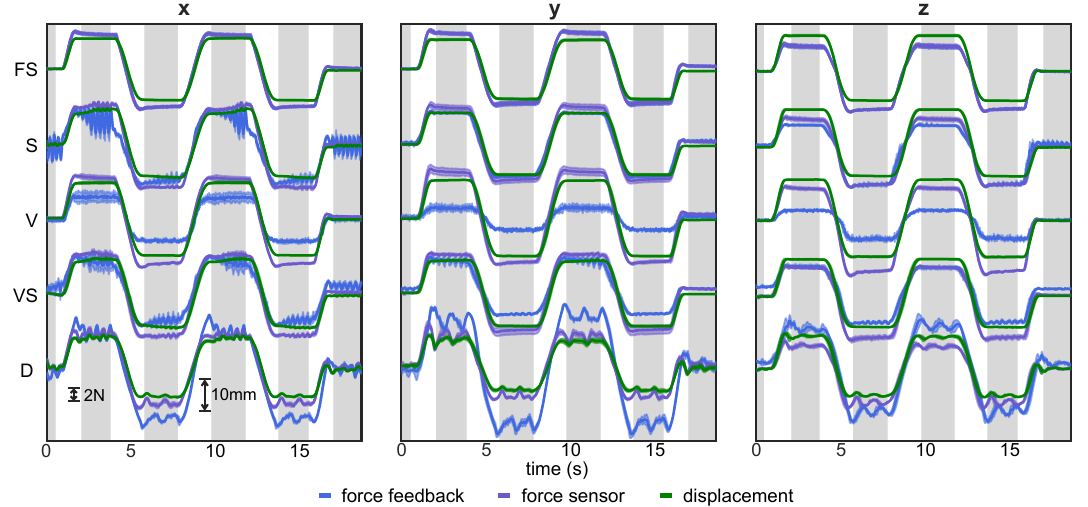}
\caption{Force feedback, ground truth force sensor measurement and displacement over time for each force estimation method. For the FS method, the force sensor measurement is also used for force feedback. Shaded regions denote $\pm1$ standard deviation. Grey areas denote the periods where the human-manipulandum model was attempting to hold a position.}
\label{figuretraj}
\end{figure*}

\subsection{Mechanical Impedance Transparency}


The fit parameters for each force estimation method  in each direction over all configurations are presented in Table\,\ref{table:stiff}. For the stiffness parameter $k$ in the x, y and z-tension directions, all neural network-based methods had lower values compared to the reference FS condition which was 302.54\,Nm\textsuperscript{-1}, 314.13\,Nm\textsuperscript{-1}, and 192.99\,Nm\textsuperscript{-1} in each direction respectively. The V network  had the largest stiffness underestimation in all directions compared to FS. The largest difference for the V network compared to FS was in the y direction, which is the direction with the least depth information available using a monocular image, where the average fitted stiffness parameter was 225.83\,Nm\textsuperscript{-1} less than that of FS. This is in contrast with the S and VS network in the y direction, where the average fitted stiffness parameter was 54.26\,Nm\textsuperscript{-1} and 71.26\,Nm\textsuperscript{-1} less than that of FS.  These observations are reflected in the stiffness plots in Fig.\,\ref{stiffplots}, where the S, V and VS networks produce less change in force over the given range of displacement in the x, y, and z-tension directions compared to the ground truth measurements shown in the darker color. In the z-palpation direction, the S and VS networks had higher stiffness values of 380.14\,Nm\textsuperscript{-1} and 367.88\,Nm\textsuperscript{-1} compared to FS at 334.82\,Nm\textsuperscript{-1}. The D force estimation model results in overestimation of stiffness in all directions compared to the FS condition. 

For the damping parameter $b$ shown in Table\,\ref{table:stiff}, the V network and D model resulted in very small or zero damping being fit. For the V network this resulted from the use of single time-frame measurements with no time history, leading to no viscoelasticity information being available to the network for force estimation. For the D model, the forces during loading and unloading exhibited inconsistencies with the viscoelastic model; there were instances where the forces in loading were lower than that in unloading, resulting in a damping parameter that would be close to or at zero given the optimization constraints. For the S network, the damping values of 39.76\,N\,m\textsuperscript{-1}, 31.66\,N\,m\textsuperscript{-1}, and 13.75\,N\,m\textsuperscript{-1}, and 24.59\,N\,m\textsuperscript{-1} are higher compared to those of the FS condition at 27.02\,N\,m\textsuperscript{-1}, 30.65\,N\,m\textsuperscript{-1}, 10.60\,N\,m\textsuperscript{-1}, 23.17\,N\,m\textsuperscript{-1} in the x, y, z-tension, and z-palpation directions respectively. This is seen in Fig.\,\ref{stiffplots} as larger hysteresis loops for the S condition in the positive and negative displacement regimes compared to the ground truth. In particular, the hysteresis loops in the x direction were large to the extent that during unloading, the force decreased rapidly to zero. For the VS network, with smaller damping values of 25.73\,N\,m\textsuperscript{-1}, 23.88\,N\,m\textsuperscript{-1}, 9.54\,N\,m\textsuperscript{-1} and 17.66\,N\,m\textsuperscript{-1} compared to the FS condition in the x, y, z-tension, and z-palpation directions respectively, the hysteresis loops were narrower, with the exception of Z3 for positive x-displacement.

For the offset parameter $c$ shown in Table\,\ref{table:stiff}, it can be seen from the increase in the standard deviations of the  parameter of the S of 0.32\,N and 0.43\,N,  and VS network of 0.87\,N and 1.00\,N in the z-tension and z-palpation directions respectively, compared to the x  and y directions (0.14\,N and 0.12\,N for S and 0.26\,N and 0.17\,N for VS in each direction respectively) that there was overfitting to the vertical position of the manipulation stage. For the S network, the height of the stage in the L1-3, C, R1-3 and M did not result in a large force offset of the force-displacement curves in Fig\,\ref{stiffplots}. However in Z3, there was a large negative offset. For VS, the height of the stage at the L1-3, C, R1-3, and M configurations produced a positive force offset in the force-displacement curves, no large force offset in Z1, and an increasing negative force offset from Z2 to Z3 (Fig.\,\ref{stiffplots}). 

 The RMSE metric in Fig.\,\ref{summary_stats}a captures the impedance transparency errors of the networks over the range of seen (L2, C, R2, Z2) and unseen (L3, L1, R1, R3, Z1, Z3, and M) configurations. For the V network, the underestimation of stiffness in the y direction is reflected in the higher RMS force error in the y direction across all configurations. For the VS network, the consistent force offset contributed to the higher RMSE compared to the S network in the z direction. If this offset is removed as shown by the dashed lines in Fig.\,\ref{summary_stats}a, the error is similar to that of S. Additionally, the overfitting to the stage height that contributed to the large standard deviation of the offset parameter $c$ resulted in the increase in RMSE from Z1 to Z3 in the VS and S networks. With the offset removed, the variation of the RMSE was reduced.
 

\subsection{Stability}
Enforcing the passivity condition on the SSM is a very conservative stabilization method that results in passivating force being applied any time the material is being unloaded. Thus, to quantify stability when the system was in an unstable state and not just temporarily active, we only considered passivating control effort during times when the human-SSM model was attempting to hold a position. The time periods were calculated by detecting movement onset and offset via the velocity profile from the unpassivated movements in the FS condition. These periods are highlighted in grey in Fig.\,\ref{figuretraj}. The RMS passivating control effort for each condition, Cartesian direction, and configuration averaged over all trials computed as shown in Fig.\,\ref{summary_stats}b. 

The force and displacement trajectories for the C configuration in Fig.\,\ref{figuretraj} show that the trajectories for S and VS in the x direction have large oscillations under closed-loop force feedback. This trend was seen in all configurations for the x direction and is commensurate with the high RMS passivating effort across all configurations for those networks compared to V (Fig.\,\ref{summary_stats}b). 

In the y direction, the S network induced in very low RMS passivating effort compared to V and VS and thus had better stability (Fig.\,\ref{summary_stats}b). The resultant force trajectories for S in the C configuration are thus shown to have small oscillations that do not induce oscillations in displacement due to the inherent damping of the modeled human-SSM (Fig.\,\ref{figuretraj}). The V network retained its stability performance as measured by RMS passivating effort in the y direction across all configurations, with the VS network matching the V network performance. This is reflected by the small force and displacement oscillations induced (Fig.\,\ref{figuretraj}). 

Likewise in the z direction,for the S network, the L3 to L1 configurations did not induce high RMS passivating effort, but there was more variability in the RMS passivating effort across C, R1 to R3, Z1 to Z3, and M (Fig.\,\ref{summary_stats}b). The V network retained a RMS passivating effort across all configurations and thus a level of stability in the z direction similar to that in the x and y directions. For the VS network, the RMS passivating effort across L1-3, C, R1-3, Z1 and Z2, was consistent but increased sharply for Z3. Given the increase in the RMS passivating effort as well as RMS force error in the z direction for the Z1-3 configurations for the S and VS networks, it is possible that the observed overfitting contributed to the degradation of stability of the networks for closed-loop force feedback.

\subsection{Demonstration of Teleoperation with Force Feedback}

\begin{figure}[!t]
\centering
\vspace{1.5mm}
\includegraphics[width=\linewidth]{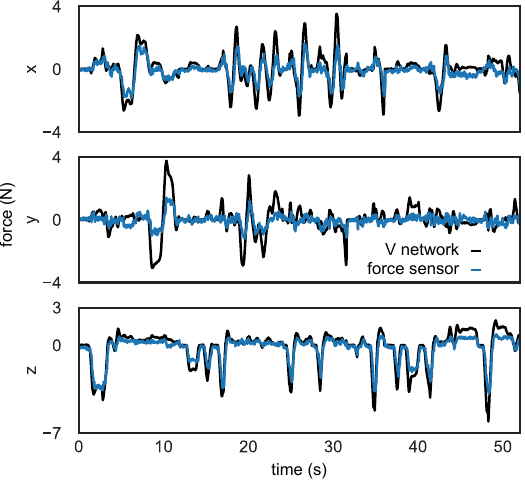}
\caption{Force feedback displayed to the human teleoperator based on estimates from the V network against the ground truth measurement from the force sensor during manipulation of artificial tissue.}
\label{demo}
\end{figure}

The network that provided the best closed-loop force feedback stability was the V network. While the S network had better stability performance as measured by RMS passivating effort in the y and z directions compared to the V network, in actual teleoperation the dynamics of the Cartesian degrees of freedom are coupled, and the instability in x caused unwanted oscillations in the other two directions. Therefore for the demonstration, we used the V network run at 60\,Hz. As shown in Fig.\,\ref{demo}, the demonstration started with slower manipulations such as a palpation in z, followed by retractions in the x and y directions within the first 12 seconds. This was followed by fast manipulations in all three Cartesian directions. At 45 seconds, there was a slow retraction in the z direction that was followed by a palpation. 

As predicted by the experiments, the impedance transparency in the y direction is low. The user did not report any unpleasant oscillations in any of the directions. The force trajectories with no visible low frequency oscillations in Fig.\,\ref{demo} support their qualitative description. Due to the poor transparency of the force feedback from V network, the environment would have felt much softer than its true stiffness. However, it is unclear if this is a significant drawback that detracts from the usefulness of the force feedback for modulating applied forces during manipulation. It is an open question whether human teleoperators benefit from exact replication of environment stiffness, or if merely presenting a variation of force in response to deformation enables the same level of performance.



\section{CONCLUSION}

Our work characterized the quality of closed-loop force feedback using neural networks during teleoperation of a dVRK in terms of impedance transparency via fitting force-displacement trajectories to a simplified viscoelastic model and stability behavior via passivity concepts. We evaluated the force feedback over various seen and unseen robot and camera configurations. We showed that a vision-only network consistently displayed poorer transparency compared to networks that relied on robot state inputs, especially in directions with low depth perception. However, such state-based networks were susceptible to large oscillatory forces that could result in instability. 

The state-only network had the best stability in two out of three of the Cartesian directions except in the configurations where there was large error from overfitting. In contrast, the vision and state-based multimodal network had stability performance that matched the vision-only networks in those directions. Thus it displayed stability performance that matched its most unstable modality in each Cartesian direction. Future work will further analyze the potential causes of this undesirable property and explore ways for a multimodal network to exploit the best stability properties from each of their input modalities.

The V network showed no tendency to induce unstable oscillations during teleoperation with force feedback despite being only run at 60Hz. This suggests that instability was not due to low update rates. This could be due to the fact that the artificial tissue had low stiffness and high damping such that the low update rate did not result in instability \cite{ColgateZwidth}. The stability at high stiffness can be explored further in use cases that contain stiff interactions like knot tying.

While the multimodal inputs and formulation of the neural network input-output as a non-dynamical system make it difficult to formulate stability guarantees, the methods presented here are shown to be capable of providing real-time force feedback that is stable in controlled and structured environments. Thus these methods can potentially be used to provide force feedback in non-safety-critical settings with consistent environments such as in surgical training. Such an application is particularly promising in the light of recent work that has explored how training in teleoperation with force information can lead to better performance in conditions where there is none \cite{GaleazziLobectomy}.

\section{ACKNOWLEDGMENTS}
The authors would like to thank Anton Deguet for his help with dVRK control, and Negin Heravi and Anthony Jarc for their advice on neural network training.


\addtolength{\textheight}{-11cm}   









\bibliographystyle{IEEEtran}
\bibliography{IEEEabrv,biblio}

\begin{thebibliography}{10}
\providecommand{\url}[1]{#1}
\csname url@samestyle\endcsname
\providecommand{\newblock}{\relax}
\providecommand{\bibinfo}[2]{#2}
\providecommand{\BIBentrySTDinterwordspacing}{\spaceskip=0pt\relax}
\providecommand{\BIBentryALTinterwordstretchfactor}{4}
\providecommand{\BIBentryALTinterwordspacing}{\spaceskip=\fontdimen2\font plus
\BIBentryALTinterwordstretchfactor\fontdimen3\font minus
  \fontdimen4\font\relax}
\providecommand{\BIBforeignlanguage}[2]{{%
\expandafter\ifx\csname l@#1\endcsname\relax
\typeout{** WARNING: IEEEtran.bst: No hyphenation pattern has been}%
\typeout{** loaded for the language `#1'. Using the pattern for}%
\typeout{** the default language instead.}%
\else
\language=\csname l@#1\endcsname
\fi
#2}}
\providecommand{\BIBdecl}{\relax}
\BIBdecl

\bibitem{goh2012global}
A.~C. Goh, D.~W. Goldfarb, J.~C. Sander, B.~J. Miles, and B.~J. Dunkin,
  ``Global evaluative assessment of robotic skills: {V}alidation of a clinical
  assessment tool to measure robotic surgical skills,'' \emph{The Journal of
  Urology}, vol. 187, no.~1, pp. 247--252, 2012.

\bibitem{martin1997objective}
J.~Martin, G.~Regehr, R.~Reznick, H.~Macrae, J.~Murnaghan, C.~Hutchison, and
  M.~Brown, ``Objective structured assessment of technical skill ({OSATS}) for
  surgical residents,'' \emph{British Journal of Surgery}, vol.~84, no.~2, pp.
  273--278, 1997.

\bibitem{Enayati2016}
N.~Enayati, E.~{De Momi}, and G.~Ferrigno, ``{Haptics in robot-assisted
  surgery: Challenges and benefits},'' \emph{IEEE Reviews in Biomedical
  Engineering}, vol.~9, pp. 49--65, 2016.

\bibitem{fontanelli2017modelling}
G.~A. Fontanelli, F.~Ficuciello, L.~Villani, and B.~Siciliano, ``Modelling and
  identification of the {da Vinci Research Kit} robotic arms,'' in
  \emph{IEEE/RSJ International Conference on Intelligent Robots and Systems},
  2017, pp. 1464--1469.

\bibitem{Pique2019dynamic}
F.~Piqu{\'e}, M.~N. Boushaki, M.~Brancadoro, E.~De~Momi, and A.~Menciassi,
  ``Dynamic modeling of the da {V}inci {R}esearch {K}it arm for the estimation
  of interaction wrench,'' in \emph{International Symposium on Medical
  Robotics}, 2019, pp. 1--7.

\bibitem{Wang2019dynamic}
Y.~Wang, R.~Gondokaryono, A.~Munawar, and G.~S. Fischer, ``A convex
  optimization-based dynamic model identification package for the da {Vinci
  Research Kit},'' \emph{IEEE Robotics and Automation Letters}, vol.~4, no.~4,
  pp. 3657--3664, 2019.

\bibitem{Haouchine2018FEA}
N.~Haouchine, W.~Kuang, S.~Cotin, and M.~Yip, ``Vision-based force feedback
  estimation for robot-assisted surgery using instrument-constrained
  biomechanical three-dimensional maps,'' \emph{IEEE Robotics and Automation
  Letters}, vol.~3, no.~3, pp. 2160--2165, 2018.

\bibitem{Tran2020}
N.~Tran, J.~Y. Wu, A.~Deguet, and P.~Kazanzides, ``A deep learning approach to
  intrinsic force sensing on the {da Vinci} surgical robot,'' in \emph{4th IEEE
  International Conference on Robotic Computing}, 2020, pp. 25--32.

\bibitem{Yilmaz2020dynamic}
N.~Yilmaz, J.~Y. Wu, P.~Kazanzides, and U.~Tumerdem, ``Neural network based
  inverse dynamics identification and external force estimation on the da
  {Vinci Research Kit},'' in \emph{IEEE International Conference on Robotics
  and Automation}, 2020, pp. 1387--1393.

\bibitem{aviles2016towards}
A.~I. Aviles, S.~M. Alsaleh, J.~K. Hahn, and A.~Casals, ``Towards retrieving
  force feedback in robotic-assisted surgery: {A} supervised
  neuro-recurrent-vision approach,'' \emph{IEEE {T}ransactions on {H}aptics},
  vol.~10, no.~3, pp. 431--443, 2016.

\bibitem{Marban2018}
A.~Marban, V.~Srinivasan, W.~Samek, J.~Fernandez, and A.~Casals, ``Estimation
  of interaction forces in robotic surgery using a semi-supervised deep neural
  network model,'' in \emph{IEEE International Conference on Intelligent Robots
  and Systems}, 2018, pp. 761--768.

\bibitem{Marban2019rnn}
A.~Marban, V.~Srinivasan, W.~Samek, J.~Fern{\'{a}}ndez, and A.~Casals, ``{A
  recurrent convolutional neural network approach for sensorless force
  estimation in robotic surgery},'' \emph{Biomedical Signal Processing and
  Control}, vol.~50, pp. 134--150, 2019.

\bibitem{LeeDaVinciNeuralNet}
Y.~E. Lee, H.~M. Husin, M.~P. Forte, S.~W. Lee, and K.~J. Kuchenbecker,
  ``Vision-based force estimation for a {da Vinci} instrument using deep neural
  networks,'' presented at the Society of American Gastrointestinal and
  Endoscoping Surgeons Meeting, 2020.

\bibitem{Shin2019transformer}
H.~Shin, H.~Cho, D.~Kim, D.~K. Ko, S.~C. Lim, and W.~Hwang, ``Sequential
  image-based attention network for inferring force estimation without haptic
  sensor,'' \emph{IEEE Access}, vol.~7, pp. 150\,237--150\,246, 2019.

\bibitem{ChuaNeuralForce}
Z.~Chua, A.~M. Jarc, and A.~M. Okamura, ``Toward force estimation in
  robot-assisted surgery using deep learning with vision and robot state,'' in
  \emph{IEEE International Conference on Robotics and Automation}, 2021, pp.
  12\,335--12\,341.

\bibitem{kazanzides2014dvrk}
P.~Kazanzides, Z.~Chen, A.~Deguet, G.~S. Fischer, R.~H. Taylor, and S.~P.
  DiMaio, ``An open-source research kit for the {da Vinci Surgical System},''
  in \emph{IEEE International Conference on Robotics and Automation}, 2014, pp.
  6434--6439.

\bibitem{deng2009imagenet}
J.~Deng, W.~Dong, R.~Socher, L.-J. Li, K.~Li, and L.~Fei-Fei, ``Imagenet: A
  large-scale hierarchical image database,'' in \emph{IEEE Conference on
  Computer Vision and Pattern Recognition}, 2009, pp. 248--255.

\bibitem{he2016resnet}
K.~He, X.~Zhang, S.~Ren, and J.~Sun, ``Deep residual learning for image
  recognition,'' in \emph{IEEE Conference on Computer Vision and Pattern
  Recognition}, 2016, pp. 770--778.

\bibitem{KingmaAdam}
D.~P. Kingma and J.~Ba, ``Adam: {A} method for stochastic optimization,''
  presented at the 3rd International Conference on Learning Representations,
  2015, arxiv:1412.6980.

\bibitem{vanbeeke2015stiffness}
F.~E. Van~Beek, W.~M.~B. Tiest, W.~Mugge, and A.~M. Kappers, ``Haptic
  perception of force magnitude and its relation to postural arm dynamics in
  {3D},'' \emph{Scientific Reports}, vol.~5, no.~1, pp. 1--11, 2015.

\bibitem{Jorda2017window}
M.~Jorda, R.~Balachandran, J.~H. Ryu, and O.~Khatib, ``{New passivity observers
  for improved robot force control},'' in \emph{IEEE International Conference
  on Intelligent Robots and Systems}, 2017, pp. 2177--2184.

\bibitem{Balachandran2017passivity}
R.~Balachandran, M.~Jorda, J.~Artigas, J.~H. Ryu, and O.~Khatib,
  ``{Passivity-based stability in explicit force control of robots},'' in
  \emph{IEEE International Conference on Robotics and Automation}, 2017, pp.
  386--393.

\bibitem{ColgateZwidth}
J.~Colgate and J.~Brown, ``Factors affecting the z-width of a haptic display,''
  in \emph{IEEE International Conference on Robotics and Automation}, 1994, pp.
  3205--3210.

\bibitem{GaleazziLobectomy}
D.~Galeazzi, A.~Mariani, S.~Sahu, S.~Maglio, E.~De~Momi, S.~Tognarelli, and
  A.~Menciassi, ``A sensorized physical simulator for training in
  robot-assisted lung lobectomy,'' presented at the Hamlyn Symposium on Medical
  Robotics, 2021.

\end{thebibliography}

\end{document}